\documentclass[runningheads]{llncs}

\usepackage{float}

\usepackage{booktabs}
\usepackage{siunitx}
\usepackage{graphicx}
\title{Towards Streaming Neural Speech Codecs through Time-Invariant Representations}
\titlerunning{Streaming Neural Speech Codecs through Time-Invariant Representations}

\author{Kélian Estève\inst{1} \and
Salima Mhdaffar\inst{1}\orcidID{0000-0002-8472-6890} \and
Mickael Rouvier\inst{1}\orcidID{0000-0003-3541-3385} \and
Richard Dufour\inst{2}\orcidID{0000-0003-1203-9108} \and
Yannick Estève\inst{1}\orcidID{0000-0002-3656-8883 }}
\authorrunning{K. Estève et al.}

\institute{LIA, Avignon Université, France \email{first.last@univ-avignon.fr}\\
\and
LS2N, Nantes Université, France
\email{first.last@univ-nantes.fr}
}

\usepackage{comment}

\begin{document}

\maketitle

\begin{abstract}

Neural speech codecs are increasingly used as intermediate representations in codec-based speech generation systems. TiCodec introduces a factorized representation that separates time-varying speech content from time-invariant information through a Time-Invariant Representation Extraction (TIRE) module, potentially reducing the amount of information that must be modeled at the frame-level.

In this work, we investigate the nature of the information captured by TIRE representations and their suitability for low-latency speech processing. Using a series of probing tasks, we analyze the influence of the encoder layer and show that intermediate layers capture complementary speaker- and environment-related information while containing little linguistic content. We further study several segment selection strategies for TIRE training and demonstrate that cross-file sampling improves the robustness of invariant representations. Based on these findings, we propose Dual-TIRE, a multi-level architecture that exploits the complementarity of different encoder layers and improves speech reconstruction quality and speaker similarity.

Finally, we evaluate TiCodec in a streaming inference setting using successive 660ms processing blocks. Results show that streaming operation can be achieved without significant degradation in reconstruction performance, highlighting the potential of factorized neural codec representations for future low-latency speech generation systems.

\end{abstract}

\section{Introduction}

Neural speech codecs have become a key component of modern speech processing systems. Originally developed for low-bitrate speech compression, neural codecs such as SoundStream~\cite{zeghidour2021soundstream}, EnCodec~\cite{defossez2022highfidelityneuralaudio}, and Descript Audio Codec (DAC)~\cite{kumar2023high} learn compact discrete representations that preserve perceptual quality while significantly reducing the bitrate of speech signals. Beyond compression, these discrete speech units have recently emerged as a versatile representation for speech generation and multimodal language modeling.

The availability of discrete speech representations has enabled a new generation of codec-based speech generation systems. Models such as AudioLM~\cite{borsos2023audiolm}, SPEAR-TTS~\cite{kharitonov2022text}, VALL-E~\cite{wang2023neural}, and Voicebox~\cite{le2023voicebox} use neural codec tokens as intermediate representations for speech synthesis and spoken language generation. By leveraging techniques initially developed for large language models, these approaches have demonstrated remarkable capabilities in speech generation, voice cloning, and spoken dialogue modeling.

Despite these advances, a major challenge remains. Neural speech codecs typically produce large numbers of frame-level tokens, often distributed across multiple Residual Vector Quantization (RVQ) layers. Consequently, speech generation systems must predict hundreds of discrete units per second, resulting in long sequences, increased computational cost, and higher inference latency. This limitation is particularly problematic for real-time speech applications, where low-latency generation is essential.

\subsection{Motivation for Streaming Speech Generation}

The challenge of efficient speech generation becomes even more critical in streaming scenarios such as spoken dialogue systems, conversational agents, simultaneous speech translation, and real-time text-to-speech synthesis. In these applications, speech must be generated continuously while only a limited amount of future context is available. The latency introduced by the prediction of large numbers of codec tokens therefore becomes a major bottleneck.

A substantial portion of the information conveyed by speech evolves slowly over time. Speaker identity, microphone characteristics, recording conditions, room acoustics, and speaking style typically remain stable throughout an utterance and often across several utterances. Nevertheless, conventional neural codecs repeatedly encode these characteristics at every frame, introducing redundancy into the representation. From the perspective of generative modeling, this redundancy forces language models to repeatedly predict information that changes little over time.

This observation motivates the development of factorized speech representations in which slowly varying information is separated from rapidly varying speech content. Such a decomposition offers two potential advantages. First, it can improve coding efficiency by avoiding the repeated encoding of invariant information. Second, it can reduce the amount of information that must be generated online, thereby facilitating low-latency and streaming speech generation. In this context, factorized neural codecs constitute a promising direction for future speech generation systems based on discrete speech units.

\subsection{Time-Invariant Representations for Neural Speech Coding}

Several recent studies have explored the explicit factorization of speech representations. FACodec, introduced as part of the NaturalSpeech~3 framework~\cite{ju2024naturalspeech}, decomposes speech into content, prosody, and timbre representations. More recently, TiCodec~\cite{ren2024fewer} proposed a neural speech codec that explicitly separates speech information into two complementary components: time-varying information and time-invariant information.

The core component of TiCodec is the Time-Invariant Representation Extraction (TIRE) module. TIRE computes a compact representation intended to capture global utterance characteristics, such as speaker identity and acoustic environment, while frame-level RVQ tokens encode local phonetic and prosodic information. During decoding, the invariant representation is combined with the temporal token sequence to reconstruct the speech signal.

This factorized representation is particularly attractive for future codec-based speech generation systems. If speaker and acoustic-context information can be represented by a small number of invariant codes, a text-to-unit model may only need to predict the temporally varying speech content. Furthermore, unlike conventional codecs that require access to the entire utterance to estimate global characteristics, such a representation may naturally support streaming operation by estimating invariant information from limited speech contexts and reusing it over multiple decoding blocks.

Despite these promising properties, several important questions remain unanswered. First, the nature of the information effectively captured by the TIRE representations remains poorly understood. While the method is intended to encode speaker and environmental characteristics, it is unclear how these properties are distributed across the different encoder layers and to what extent linguistic information is removed from the invariant representation. Second, the influence of the segment selection strategy used during TIRE training has not been systematically investigated. Finally, although the architecture appears particularly well suited to streaming speech generation, its behavior under streaming inference conditions has not yet been studied.

In this paper, we address these questions through a comprehensive analysis of the TIRE module. We first investigate the information encoded by time-invariant representations using a series of probing tasks covering speaker identification, acoustic scene classification, emotion recognition, language identification, and keyword spotting. We then study the influence of the encoder layer and segment selection strategy on the quality of the learned representations. Based on the observed complementarity between different representation levels, we introduce a Dual-TIRE architecture that exploits invariant information extracted at multiple encoder layers. Finally, we evaluate TiCodec in a streaming inference configuration based on successive 660,ms processing blocks and demonstrate that the codec maintains reconstruction quality without requiring access to the complete utterance.

The contributions of this work are summarized as follows:
\begin{itemize}
\item We provide a detailed characterization of the information captured by TiCodec time-invariant representations through probing analyses covering speaker, acoustic, paralinguistic, and linguistic attributes.

\item We investigate the influence of extraction layers and segment selection strategies and propose a Dual-TIRE architecture that exploits complementary invariant information across representation levels.

\item We demonstrate that TiCodec can operate in a streaming inference regime using 660ms processing blocks while maintaining reconstruction performance, highlighting the potential of factorized neural codec representations for low-latency speech generation systems.
\end{itemize}

\section{Overview of the TiCodec and Dual-TIRE Architectures}
\label{architecture}

We first present the original TiCodec architecture and its TIRE module (Section~\ref{ticodec_architecture}), and then introduce the Dual-TIRE architecture proposed in this paper (Section~\ref{dual-tire_architecture_overview}).

\subsection{TiCodec Architecture}
\label{ticodec_architecture}

TiCodec~\cite{ren2024fewer} is a neural audio codec based on the principle of factorizing speech information into two complementary components: (i) time-dependent information, associated with local signal variations (phonetic and prosodic content), and (ii) time-invariant information, corresponding to global utterance attributes such as speaker identity or acoustic scene characteristics. This explicit separation aims to improve coding efficiency by avoiding redundancy in the representation of global information.

\begin{figure}[ht] 
    \centering 
    \includegraphics[width=0.7\columnwidth]{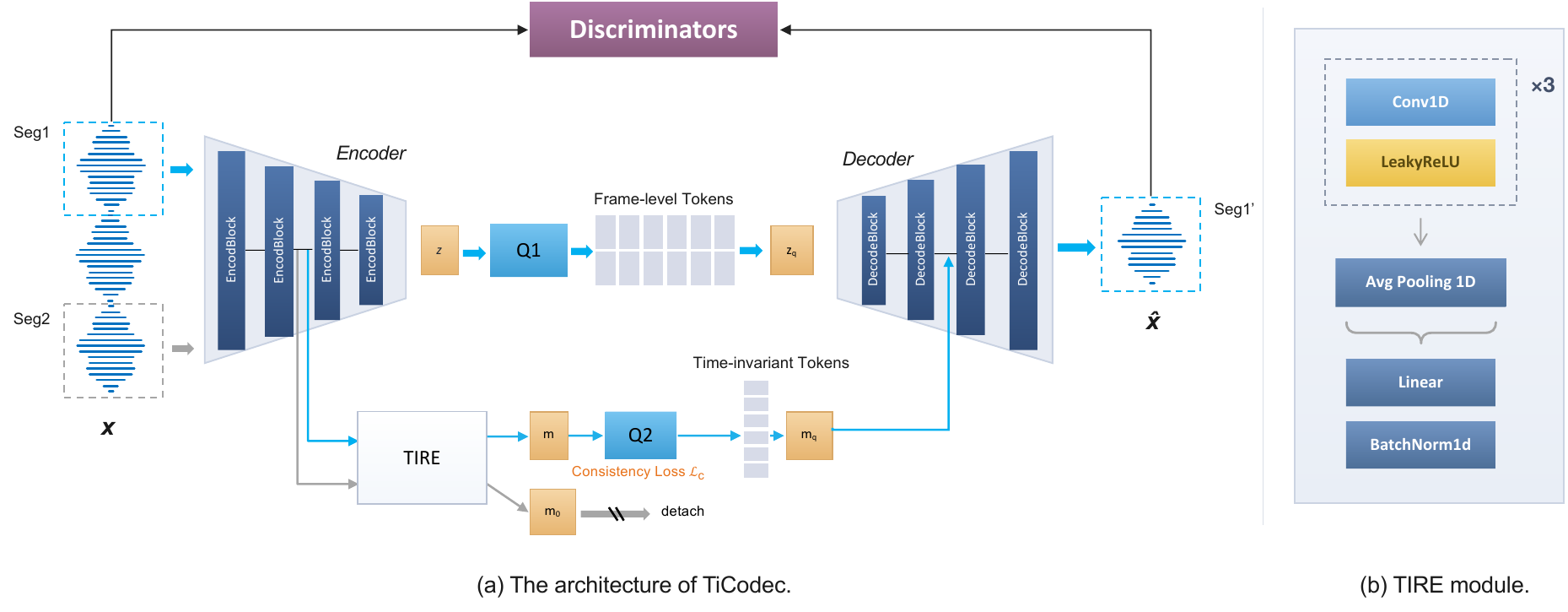} 

    \caption{TiCodec architecture, illustrating the factorization of speech into time-dependent representations encoded via RVQ and time-invariant representations extracted by the TIRE module}
    \label{fig:ticodec} 
\end{figure}

Figure~\ref{fig:ticodec} illustrates the general architecture of TiCodec. The framework is built upon a convolutional autoencoder. The encoder transforms the input audio into a sequence of latent representations. These representations are subsequently quantized using a RVQ mechanism, enabling efficient discretization of phonetic and prosodic information at the frame-level. TiCodec incorporates a module named TIRE, which is responsible for extracting time-invariant information. It is designed to capture stationary utterance characteristics, such as speaker identity and global acoustic environment properties, that do not require fine-grained temporal modeling. During the decoding phase, the audio signal is reconstructed by combining the temporal tokens derived from the RVQ and the invariant tokens produced by TIRE. The latter are replicated along the time-axis to provide the decoder with consistent global conditioning, thereby contributing to the preservation of speaker timbre and acoustic characteristics.

\subsection{Proposed Dual-TIRE Architecture}
\label{dual-tire_architecture_overview}

The original TiCodec architecture relies on a single TIRE module connected to one encoder layer and reinjected into the corresponding decoder layer. This configuration assumes that a single representation level is sufficient to capture the invariant information required for reconstruction. However, we hypothesize that different encoder depths may provide partially complementary information: lower layers are expected to encode more acoustic-oriented cues, while higher layers may capture more abstract, high-level information useful for conditioning the reconstruction process.

\begin{figure}[ht] 
    \centering 
    \includegraphics[width=0.7\columnwidth]{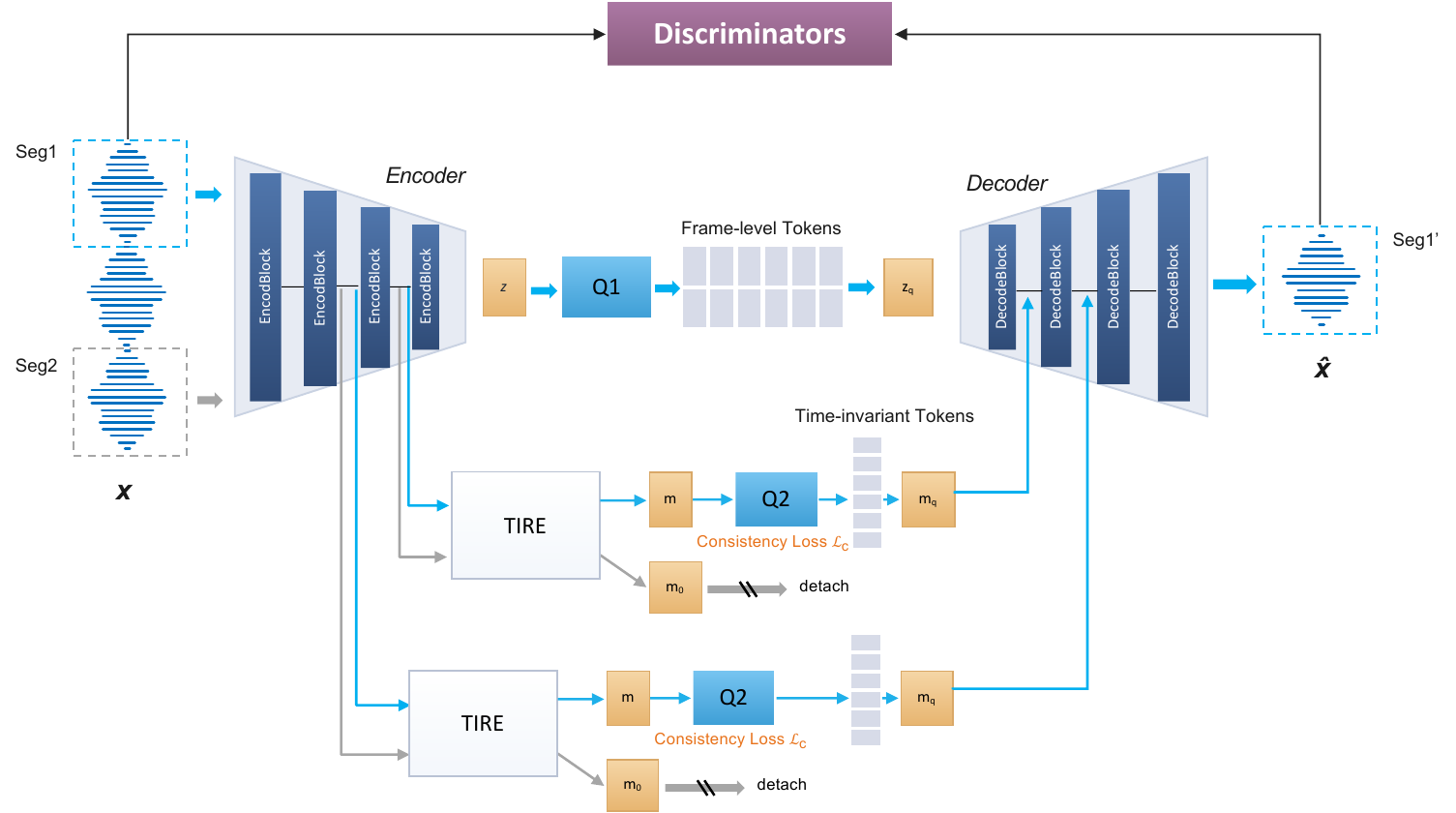}

    \caption{Dual-TIRE architecture. Two TIRE modules extract time-invariant representations from encoder layers 2 and 3, which are quantized independently and reinjected into the corresponding decoder layers, while frame-level content is encoded via RVQ.}
    \label{fig:dual-tire}    
\end{figure}

We therefore propose Dual-TIRE, a multi-level extension of TiCodec that integrates two independent TIRE modules. As illustrated in Figure~\ref{fig:dual-tire}, each module extracts invariant representations from a distinct encoder level. Both streams are quantized independently and reinjected into their corresponding decoder layers. We refer to the \textit{connection layer} as the symmetric layer index at which a TIRE branch extracts features from the encoder and reinjects the resulting invariant representation into the decoder.

This dual extraction and injection mechanism is designed to enrich the global conditioning available to the decoder by combining invariant representations extracted at different hierarchical levels. The frame-level RVQ stream remains unchanged, while the additional TIRE branch introduces a second global conditioning signal.

\section{Protocol and Probing Tasks}
\label{protocol}

To analyze the information actually captured by the TIRE representations, we adopt a diagnostic probing approach that consists of evaluating their capacity to encode the information necessary for a set of targeted classification tasks. The underlying premise is that strong performance on a given task indicates the presence of the relevant information, while weak performance suggests that the information is absent or significantly attenuated.


To evaluate the intrinsic quality of the extracted features, the representations produced by the TIRE module are kept frozen and used as inputs to a downstream diagnostic classifier. This classifier is a Multi-Layer Perceptron (MLP) composed of two fully-connected hidden layers of 1,024 neurons each, followed by an output layer adapted to the task. All experiments are evaluated using accuracy. By keeping the codec frozen and avoiding joint fine-tuning, this protocol ensures that we explicitly measure the information already captured and preserved by the TIRE representations.


To cover a wide range of information likely to be encoded in TIRE representations, we consider five representative classification tasks, ranging from speaker identity to linguistic and semantic content.

\noindent \textbf{Speaker Identification:} This task aims to determine whether an acoustic representation contains sufficient discriminative information to identify a speaker by their voice. In this framework, the problem is formulated as a multi-class classification, where each class corresponds to a distinct speaker. TIRE representations are used to train a classifier on 1,022 speakers from the VoxCeleb1 dataset~\cite{nagrani2020voxceleb}.

\noindent \textbf{Acoustic Scene Classification:} This task evaluates the capacity of TIRE representations to capture information related to the acoustic environment. We use the TAU Urban Acoustic Scenes corpus~\cite{mesaros2018multi}, composed of 10 classes corresponding to different urban sound contexts (Airport, Bus, Metro, Park, etc.).

\noindent \textbf{Emotion Recognition:} Emotion recognition aims to analyze whether representations derived from TiCodec preserve paralinguistic parameters, such as intonation, energy, or speech rhythm. This task is performed on the MELD corpus~\cite{poria2019meld}, which provides annotations for 7 emotional classes.

\noindent \textbf{Spoken Language Identification:} This task measures the capacity of representations to encode high-level linguistic information, particularly the phonetic and prosodic characteristics specific to a language. We use a multilingual version of the Common Voice corpus~\cite{ardila2020common}, covering 45 languages. High accuracy in this task would indicate that TIRE representations retain sufficient parameters to differentiate languages, despite the compression process performed by the neural codec.

\noindent \textbf{Keyword Spotting:} Finally, the keyword spotting task aims to evaluate the preservation of local lexical information. We use the Google Speech Commands corpus~\cite{warden2018speech}, composed of 35 classes corresponding to different keywords (Yes, No, Up, Down, Left, Right, etc.).

\section{Experiments and Results}
\label{experiments}

This section presents the experimental setup and results. It introduces the datasets and evaluation metrics (Section~\ref{experimental_protocol}), then analyzes the standard TiCodec architecture: the impact of the TIRE connection layer together with a probing characterization of the encoded information (Section~\ref{analysis_TIRE_module}), and the segment selection strategies used during TIRE training (Section~\ref{analysis_segment_strategies}). Building on these analyses, we finally evaluate the proposed Dual-TIRE architecture (Section~\ref{dual-tire_architecture}).


\subsection{Experimental Protocol}
\label{experimental_protocol}

We first describe the speech corpora used for training and evaluation, followed by the objective metrics adopted to assess speech quality, intelligibility, spectral fidelity, and speaker identity preservation.

\noindent \textbf{Training:} The TiCodec model is trained on the LibriTTS corpus~\cite{zen2019libritts}, a multi-speaker English read speech dataset recorded at a sampling rate of 24 kHz. LibriTTS is derived from the materials used to create LibriSpeech~\cite{panayotov2015librispeech}, combining audio recordings from LibriVox and text transcriptions from Project Gutenberg. The corpus totals approximately 585 hours of speech, produced by 2,456 speakers. For training, we use the train-clean-100, train-clean-360, and train-other-500 subsets.

\noindent \textbf{Evaluation:} The evaluation of the different models is performed on the LibriTTS test-clean~\cite{zen2019libritts}, VCTK~\cite{veaux2017cstr} and EMILIA~\cite{he2024emilia}. For EMILIA, 4,000 utterances are randomly selected for each language: French, German, Japanese and Korean.

\noindent \textbf{Metrics:} Speech quality and intelligibility assessment was performed using the following metrics: ViSQOL V3 (Virtual Speech Quality Objective Listener)~\cite{chinen2020visqolv3opensource}, PESQ (Perceptual Evaluation of Speech Quality)~\cite{941023}, STOI (Short-Time Objective Intelligibility)~\cite{5713237}, and Mel Cepstral Distortion (MCD)~\cite{407206}. Speaker identity preservation is evaluated using Sim, computed as the cosine similarity between speaker embeddings extracted from the original and reconstructed utterances using a pretrained speaker-verification model. Higher values indicate better preservation of speaker identity. All these metrics allow for a joint evaluation of global perceptual quality, intelligibility, spectral fidelity, as well as the preservation of speaker characteristics. Together, they provide a comprehensive assessment of system performance.

\subsection{Analysis of the Impact of the TIRE Module Connection Layer}
\label{analysis_TIRE_module}


To investigate the informative potential of other layers within the architecture, we trained several distinct systems in which the TIRE module is connected to layers 1, 2, 3, and 4, respectively, in a symmetric manner across the encoder and decoder. The objective of this study is to identify the layer yielding the richest representations while limiting redundancy with time-dependent information.

Table~\ref{resultat_analayse_couche_couche} reports the results, on the LibriTTS test-clean corpus, obtained for a reference system without the TIRE module (\textit{No-TIRE}), as well as for the different variants where TIRE is connected to the corresponding encoder and decoder layers.

\begin{table}[H]
\centering
\caption{Impact of the TIRE injection layer on LibriTTS test-clean. Intermediate layers (2 and 3) provide the best trade-off between perceptual quality, intelligibility, spectral fidelity, and speaker similarity. Best results are shown in bold.}

\begin{tabular}{lccccc}
\toprule
\textbf{TIRE injection layer} & \textbf{ViSQOL}$\uparrow$ & \textbf{PESQ}$\uparrow$ & \textbf{STOI}$\uparrow$ & \textbf{MCD}$\downarrow$ & \textbf{Sim}$\uparrow$ \\
\midrule
No-TIRE (baseline) & 4.362 & 2.797 & 0.938 & 0.772 & 0.703 \\
\midrule
Layer~1 & 4.373 & 2.897 & 0.942 & 0.742 & 0.721 \\
Layer~2 & 4.382 & \textbf{2.956} & \textbf{0.944} & \textbf{0.728} & 0.699 \\
Layer~3 & \textbf{4.406} & 2.876 & 0.942 & 0.762 & \textbf{0.722} \\
Layer~4 & 4.328 & 2.748 & 0.935 & 0.794 & 0.696 \\
\bottomrule
\end{tabular}

\label{resultat_analayse_couche_couche}

\end{table}

Analysis of Table~\ref{resultat_analayse_couche_couche} indicates that TIRE generally improves metrics over the No-TIRE baseline, except for Layer~4. However, improvements are not uniform across all criteria, as speaker similarity does not always increase. Among the various configurations, the systems utilizing Layers 2 and 3 appear to offer the best trade-off. Specifically, the system using Layer 2 optimizes signal sharpness and fine structure, achieving a PESQ score of 2.956 and an MCD of 0.728. Conversely, the system trained on Layer 3 favors superior global perceptual fidelity and better preservation of speaker identity, as evidenced by a ViSQOL score of 4.406 and a Sim score of 0.722.



To better understand the nature of the information driving these performance differences, we evaluated the frozen TIRE representations from each encoder layer across the five selected downstream probing classification tasks (see Section~\ref{protocol}).


\begin{figure}[ht]

    \centering 
    \includegraphics[width=\columnwidth]{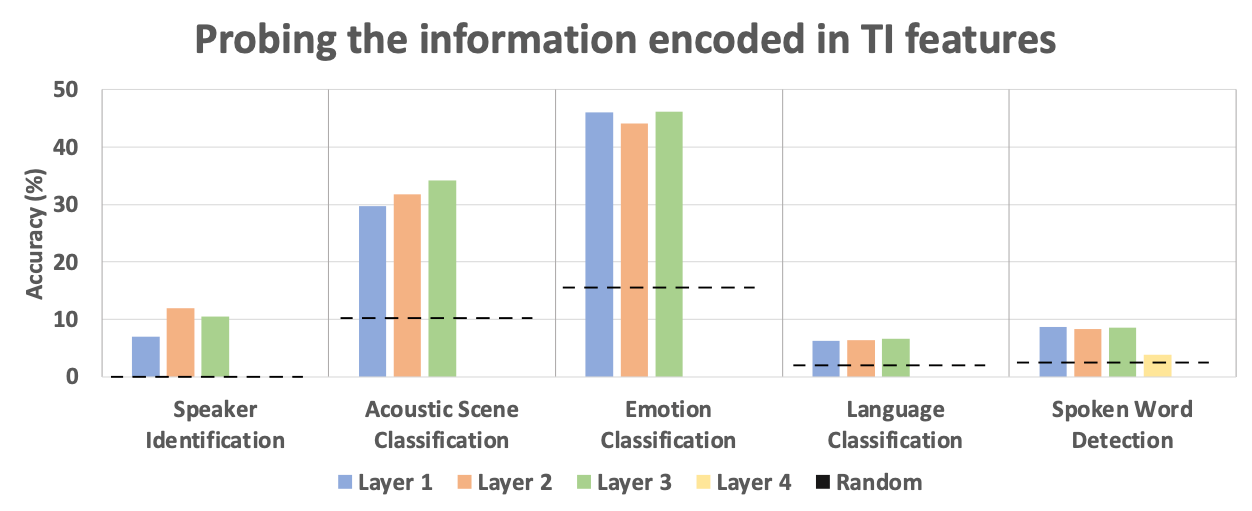} 
    
    \caption{Probing evaluation of TIRE representations extracted from encoder Layers 1–4 across five downstream classification tasks, highlighting the types of information captured at each layer and the progressive loss of informative content in deeper compressed representations.}

    \label{fig:probing} 
\end{figure}

The results presented in Figure~\ref{fig:probing} indicate that Layers 1, 2, and 3 generally capture information of a similar nature. We observe that the TIRE module encodes parameters related to acoustic scenes and emotions, suggesting effective capture of global and paralinguistic signal attributes. Conversely, performance on spoken language identification and keyword spotting tasks is near chance level, indicating that the TIRE module does not retain fine-grained linguistic or lexical information.

Regarding speaker identity, the results remain moderate. Although some discriminative information is present, the relatively low performance suggests that the TIRE module does not explicitly encode speaker identity, but rather high-level acoustic features likely correlated with it.

Finally, a marked degradation in performance is observed for Layer 4, regardless of the task considered. This layer does not appear to encode information exploitable by the TIRE module, likely because the resulting representations are too compressed.

\subsection{Analysis of Segment Selection Strategies for TIRE Module Training}
\label{analysis_segment_strategies}

During the training of TiCodec, and specifically the TIRE module, two audio segments are extracted from the audio file to constrain the learning of invariant representations. In the standard configuration, these two segments are drawn randomly from the same audio file, without any explicit constraint on their temporal positions.

The underlying principle is to encourage TIRE to produce similar representations for segments sharing common global attributes (speaker identity, acoustic scene), while reducing the influence of time-dependent information, such as phonetic or lexical content. In other words, the TIRE module is incentivized to minimize representation variations between two segments originating from the same source, regardless of their local differences, in order to foster the encoding of invariant information.

However, the manner in which these segments are selected during training can have a direct impact on the nature of the information effectively learned. In this section, we propose five distinct segment selection strategies. Each strategy defines a specific sampling policy for the two segments used during TIRE training:

\begin{itemize}

\item\textbf{Random segments (unconstrained):} Two segments are extracted randomly from the same audio file, without any constraints on their temporal positions. The segments may overlap. This strategy corresponds to the default configuration and serves as a baseline.

\item\textbf{Random non-overlapping segments:} Two segments are extracted randomly from the same file, with the additional constraint that they do not overlap in time. This strategy aims to reduce direct redundancy between segments while retaining an identical global context.

\item\textbf{Adjacent segments:} The first segment is extracted randomly from the file, and the second segment corresponds to the contiguous segment immediately following the first. This strategy introduces strong temporal correlation between segments and favors local signal continuity.

\item\textbf{Cross-file (same speaker):} The first segment is extracted from the current file, while the second is extracted from a different file belonging to the same speaker. By exploiting recordings from different sessions of the same speaker, this strategy allows TIRE to build a representation space that better accounts for speaker variability (e.g., changes in channel, recording conditions, or speaking style).

\item\textbf{Cross-file (same speaker and book)}: The first segment is extracted from the current file, and the second from a different file recorded by the same speaker and originating from the same book. This strategy reduces speaker variability.

\end{itemize}

The results presented in Table~\ref{segment_couche_2} show that the segment extraction strategy influences performance differently depending on the Layer 2 and 3. When representations are extracted at Layer 2, the unconstrained strategy (random segments) offers the best performance across all metrics, suggesting that free extraction favors a good trade-off between reconstruction quality and speaker identity preservation.

\begin{table}[H]
\centering
\caption{Impact of segment selection strategies during TIRE training on LibriTTS test-clean. Representations are extracted from encoder Layer~2 and Layer~3. Best results within each Layer are shown in bold.}
\begin{tabular}{lccccc}
\toprule
\textbf{Segment selection strategy / Layer} & \textbf{ViSQOL}$\uparrow$ & \textbf{PESQ}$\uparrow$ & \textbf{STOI}$\uparrow$ & \textbf{MCD}$\downarrow$ & \textbf{Sim}$\uparrow$ \\
\midrule
\multicolumn{6}{c}{\textit{Layer~2}} \\
\midrule
Random segments (unconstrained)      & \textbf{4.382} & \textbf{2.956} & \textbf{0.944} & \textbf{0.728} & \textbf{0.699} \\
Random non-overlapping segments      & 4.355 & 2.827 & 0.940 & 0.763 & 0.691 \\
Adjacent segments                    & 4.373 & 2.901 & 0.944 & 0.744 & 0.696 \\
Cross-file (same speaker)            & 4.363 & 2.859 & 0.942 & 0.749 & 0.687 \\
Cross-file (same speaker, same book) & 4.361 & 2.864 & 0.942 & 0.746 & 0.683 \\
\midrule
\midrule
\multicolumn{6}{c}{\textit{Layer~3}} \\
\midrule
Random segments (unconstrained) & 4.406 & 2.876 & 0.942 & 0.762 & 0.722 \\
Random non-overlapping segments  & 4.382 & 2.865 & 0.943 & 0.761 & 0.722 \\
Adjacent segments & 4.392 & 2.860 & 0.942 & 0.779 & 0.724 \\ 
Cross-file (same speaker) & 4.399 & 2.859 & \textbf{0.943} & 0.755 & \textbf{0.730} \\ 
Cross-file (same speaker, same book) & \textbf{4.409} & \textbf{2.882} & 0.942 & \textbf{0.754} & 0.724 \\

\bottomrule
\end{tabular}

\label{segment_couche_2}
\end{table}

In contrast, at Layer 3, cross-file strategies, particularly the one relying on distinct files from the same speaker and the same book, lead to slight improvements in terms of perceptual fidelity and speaker similarity. This indicates that at deeper representation layers, exposing TIRE to controlled inter-utterance variability helps reinforce the encoding of invariant attributes.

\subsection{Dual-TIRE Architecture}
\label{dual-tire_architecture}

We now evaluate whether the complementarity observed between Layers~2 and~3 translates into improved reconstruction performance. We compare Dual-TIRE with the No-TIRE and TIRE baselines.

Furthermore, two variants of the Dual-TIRE architecture are evaluated. They differ in the segment selection strategy employed during training:
In the Dual-TIRE (baseline) configuration, both TIRE modules are trained using the random (unconstrained) segment selection strategy.
Conversely, the Dual-TIRE (cross-file) configuration adopts a differentiated training strategy depending on the layer level. The TIRE module connected to layer 2 is trained using the baseline strategy, whereas the TIRE module connected to layer 3 is trained using a cross-file (same speaker and book) strategy.

Table~\ref{tab:comparison_dual_tire} reports the results on LibriTTS test-clean. The TIRE system improves over the No-TIRE baseline on most reconstruction-oriented metrics, notably PESQ and MCD, confirming the benefit of injecting time-invariant information into the decoder. The Dual-TIRE variants further modify this trade-off. The Dual-TIRE (baseline) configuration slightly improves ViSQOL, STOI, and MCD compared with the single-TIRE system, reaching the best MCD value of 0.727 and the best STOI value of 0.946. However, PESQ remains lower than that of the TIRE baseline.

The Dual-TIRE (cross-file) variant provides the best ViSQOL score, with 4.410, and the highest speaker similarity, with 0.721. This suggests that applying a layer-dependent segment selection strategy can strengthen the preservation of global attributes, especially speaker-related information. Nevertheless, this variant does not uniformly improve all metrics: PESQ decreases compared with single-TIRE, and MCD is slightly worse than both TIRE and Dual-TIRE (baseline). These results indicate that Dual-TIRE introduces a different reconstruction trade-off rather than a systematic improvement across all objective criteria. Importantly, this trade-off comes at a negligible cost: the second TIRE branch adds only 0.57\,M parameters (+0.9\%) over the single-TIRE system.

\begin{table}[H]
\centering
\caption{Comparison between No-TIRE, TIRE, and Dual-TIRE architectures on the LibriTTS test-clean dataset. Params denotes the total number of model parameters.}
\begin{tabular}{lcccccc}
\toprule
\textbf{System} & \textbf{Params} & \textbf{ViSQOL}$\uparrow$ & \textbf{PESQ}$\uparrow$ & \textbf{STOI}$\uparrow$ & \textbf{MCD}$\downarrow$ & \textbf{Sim}$\uparrow$ \\
\midrule
No-TIRE (baseline) & 63.12\,M &  4.362 & 2.797 & 0.938 & 0.772 & 0.703 \\
\midrule
TIRE & 63.33\,M & 4.382 & \textbf{2.956} & 0.944 & 0.728 & 0.699 \\
Dual-TIRE (baseline)   & 63.91\,M & 4.387 & 2.931 & \textbf{0.946} & \textbf{0.727} & 0.699 \\
Dual-TIRE (cross-file) & 63.91\,M  & \textbf{4.410} & 2.923 & \textbf{0.946} & 0.741 & \textbf{0.721} \\
\bottomrule
\end{tabular}
\label{tab:comparison_dual_tire}
\end{table}

We further evaluate the generalization ability of Dual-TIRE on out-of-domain datasets, including VCTK, and multilingual EMILIA subsets. The results are reported in Table~\ref{tab:out_of_domain_comparison}. Compared with TIRE, Dual-TIRE improves the average ViSQOL score from 4.307 to 4.318 and increases the average speaker similarity from 0.681 to 0.701. These gains are particularly consistent for speaker similarity, where Dual-TIRE outperforms TIRE on every evaluated corpus.

However, the improvement is not uniform across all metrics. PESQ decreases on all out-of-domain datasets, and STOI remains unchanged on average. MCD is also very close on average, with mixed behavior depending on the corpus. These results suggest that Dual-TIRE mainly benefits perceptual quality and speaker identity preservation, while preserving intelligibility and maintaining comparable spectral distortion. The decrease in PESQ indicates a trade-off between global conditioning and some reconstruction-oriented measures.

Overall, these findings support the hypothesis that Layers~2 and~3 provide complementary invariant information. The joint use of both levels appears particularly useful for preserving speaker-related and perceptual attributes in out-of-domain conditions. At the same time, the relatively small differences between systems highlight the need to interpret Dual-TIRE as a trade-off between reconstruction quality, speaker identity preservation, and additional invariant-code capacity.


\begin{table}[H]
\centering
\caption{Comparison between TIRE and Dual-TIRE architectures on out-of-domain datasets. Best results within each corpus are shown in bold.}

\begin{tabular}{lccccc}

\toprule
\textbf{Dataset / System} & \textbf{ViSQOL}$\uparrow$ & \textbf{PESQ}$\uparrow$ & \textbf{STOI}$\uparrow$ & \textbf{MCD}$\downarrow$ & \textbf{Sim}$\uparrow$ \\
\midrule
\multicolumn{6}{c}{\textit{TIRE}} \\
\midrule
VCTK        & 4.330 & \textbf{2.598} & 0.851 & \textbf{0.624} & 0.660 \\
EMILIA-DE   & \textbf{4.382} & \textbf{2.956} & \textbf{0.944} & 1.728 & 0.701 \\
EMILIA-FR   & 4.281 & \textbf{2.429} & 0.918 & \textbf{1.586} & 0.681 \\
EMILIA-JA   & 4.304 & \textbf{2.534} & 0.923 & \textbf{1.535} & 0.689 \\
EMILIA-KO   & 4.238 & \textbf{2.350} & 0.915 & \textbf{1.805} & 0.674 \\
\midrule
\textbf{Average} & 4.307 & \textbf{2.573} & \textbf{0.910} & 1.456 & 0.681 \\
\midrule
\midrule
\multicolumn{6}{c}{\textit{Dual-TIRE}} \\
\midrule
VCTK        & \textbf{4.346} & 2.434 & \textbf{0.859} & 0.629 & \textbf{0.670} \\
EMILIA-DE   & 4.317 & 2.341 & 0.927 & \textbf{1.581} & \textbf{0.724} \\
EMILIA-FR   & \textbf{4.313} & 2.392 & \textbf{0.920} & 1.630 & \textbf{0.702} \\
EMILIA-JA   & \textbf{4.335} & 2.504 & \textbf{0.925} & 1.573 & \textbf{0.711} \\
EMILIA-KO   & \textbf{4.281} & 2.319 & \textbf{0.918} & 1.864 & \textbf{0.697} \\
\midrule
\textbf{Average} & \textbf{4.318} & 2.398 & \textbf{0.910} & \textbf{1.455} & \textbf{0.701} \\
\bottomrule
\end{tabular}

\label{tab:out_of_domain_comparison}

\end{table}

\section{Streaming capabilities of TiCodec}
Since the TiCodec neural codec was originally trained on short audio segments of 660 ms, its ability to generalize to longer utterances was examined under two different decoding conditions. 

In the offline setting, the model was provided with entire sequences at once to evaluate how well it could maintain coherence beyond its training context. In the streaming setting, audio was processed block by block to simulate real-time usage, where only the current segment is available for reconstruction. Each block lasts 660ms, the same duration as during the training process. Blocks are contiguous, no overlap has been introduced.
Each segment generated by TiCodec is directly concatenated with the previously reconstructed signal, without any smoothing.

To evaluate the streaming capabilities of TiCodec, we report three complementary metrics. Mean Opinion Score (MOS) estimates the perceptual quality of reconstructed speech and reflects the subjective judgment of human listeners. Higher MOS values indicate better perceived speech quality. Scale-Invariant Signal-to-Distortion Ratio (SI-SDR) measures the fidelity of the reconstructed signal with respect to the reference signal while being insensitive to global amplitude scaling. It quantifies the amount of distortion introduced during reconstruction, with higher values corresponding to better signal preservation. Finally, Weighted Signal-to-Noise Ratio (WSNR) evaluates the ratio between the reference signal and the reconstruction error after perceptual weighting, providing an estimate of the perceptually relevant distortion. As with SI-SDR, larger WSNR values indicate higher reconstruction quality.

Table~\ref{tab:comparison_offline_streaming} presents the results of the comparison between offline and streaming mode by using different metrics. 
For this experiment, we used the TiCodec model trained on LibriTTS and made the comparison on the LibriTTS validation dataset.

This comparison highlights the trade-off between context-rich reconstructions in the offline case and the low-latency, block-wise decoding required for interactive applications, shedding light on TiCodec’s potential for both post-processing and real-time deployment. 
Indeed, the results indicate that streaming operation can be achieved without significant performance degradation.

\begin{table}[H]
\caption{Comparison of the performance when the inference is produced offline (i.e. by processing the entire audio file at once) vs. online.}
\centering
\begin{tabular}{lccccc}
\toprule
\textbf{Inference mode} & \textbf{MOS}$\uparrow$ & \textbf{PESQ}$\uparrow$ & \textbf{SISDR}$\uparrow$ & \textbf{STOI}$\uparrow$ & \textbf{WSNR}$\uparrow$ \\
\midrule
Offline   & 0.621 & 0.767 & 0.751 & 0.992 & 0.276 \\
Streaming & 0.618 & 0.728 & 0.770 & 0.991 & 0.272 \\
\bottomrule
\end{tabular}

\label{tab:comparison_offline_streaming}
\end{table}

\section{Conclusion}

This work investigated the role of time-invariant representations in the TiCodec neural speech codec and their potential for low-latency speech processing. Through probing analyses, we showed that the TIRE module primarily captures global acoustic and paralinguistic information while retaining little linguistic content. We further demonstrated that the choice of extraction layer and segment selection strategy significantly influences the quality and robustness of the learned invariant representations.

Building on these observations, we proposed Dual-TIRE, a multi-level architecture that exploits the complementarity of invariant representations extracted from different encoder layers. Experimental results on both in-domain and out-of-domain datasets showed consistent improvements in speech quality and speaker similarity compared to the original TiCodec architecture.

Finally, we evaluated TiCodec in a streaming inference setting based on successive 660\,ms processing blocks. The results indicate that streaming operation can be achieved with only minor performance degradation, confirming that factorized neural codec representations constitute a promising direction for future low-latency speech generation systems.

\section*{Acknowledgement}

This work was supported by 2022 Jelinek Memorial Summer Workshop on Speech and Language Technologies  at BUT@Brno. This project has also received funding from the European Union’s Horizon 2020 research and innovation programme under the Marie Skłodowska-Curie grant agreement No 101007666.
This work was performed using HPC resources from GENCI–IDRIS (AD011012551R4 and AD011012108R3).

\bibliographystyle{IEEEtran}
\bibliography{mybib}

\begin{thebibliography}{10}
\providecommand{\url}[1]{#1}
\csname url@samestyle\endcsname
\providecommand{\newblock}{\relax}
\providecommand{\bibinfo}[2]{#2}
\providecommand{\BIBentrySTDinterwordspacing}{\spaceskip=0pt\relax}
\providecommand{\BIBentryALTinterwordstretchfactor}{4}
\providecommand{\BIBentryALTinterwordspacing}{\spaceskip=\fontdimen2\font plus
\BIBentryALTinterwordstretchfactor\fontdimen3\font minus \fontdimen4\font\relax}
\providecommand{\BIBforeignlanguage}[2]{{%
\expandafter\ifx\csname l@#1\endcsname\relax
\typeout{** WARNING: IEEEtran.bst: No hyphenation pattern has been}%
\typeout{** loaded for the language `#1'. Using the pattern for}%
\typeout{** the default language instead.}%
\else
\language=\csname l@#1\endcsname
\fi
#2}}
\providecommand{\BIBdecl}{\relax}
\BIBdecl

\bibitem{zeghidour2021soundstream}
N.~Zeghidour, A.~Luebs, A.~Omran, J.~Skoglund, and M.~Tagliasacchi, ``Soundstream: An end-to-end neural audio codec,'' \emph{IEEE/ACM Transactions on Audio, Speech, and Language Processing}, vol.~30, pp. 495--507, 2021.

\bibitem{defossez2022highfidelityneuralaudio}
\BIBentryALTinterwordspacing
A.~Défossez, J.~Copet, G.~Synnaeve, and Y.~Adi, ``High fidelity neural audio compression,'' 2022. [Online]. Available: \url{https://arxiv.org/abs/2210.13438}
\BIBentrySTDinterwordspacing

\bibitem{kumar2023high}
R.~Kumar, P.~Seetharaman, A.~Luebs, I.~Kumar, and K.~Kumar, ``High-fidelity audio compression with improved rvqgan,'' \emph{Advances in Neural Information Processing Systems}, vol.~36, pp. 27\,980--27\,993, 2023.

\bibitem{borsos2023audiolm}
Z.~Borsos, R.~Marinier, D.~Vincent, E.~Kharitonov, O.~Pietquin, M.~Sharifi, D.~Roblek, O.~Teboul, D.~Grangier, M.~Tagliasacchi \emph{et~al.}, ``Audiolm: a language modeling approach to audio generation,'' \emph{IEEE/ACM transactions on audio, speech, and language processing}, vol.~31, pp. 2523--2533, 2023.

\bibitem{kharitonov2022text}
E.~Kharitonov, A.~Lee, A.~Polyak, Y.~Adi, J.~Copet, K.~Lakhotia, T.-A. Nguyen, M.~Riviere, A.~Mohamed, E.~Dupoux \emph{et~al.}, ``Text-free prosody-aware generative spoken language modeling,'' in \emph{ACL}, 2022.

\bibitem{wang2023neural}
C.~Wang, S.~Chen, Y.~Wu, Z.~Zhang, L.~Zhou, S.~Liu, Z.~Chen, Y.~Liu, H.~Wang, J.~Li \emph{et~al.}, ``Neural codec language models are zero-shot text to speech synthesizers,'' \emph{arXiv preprint arXiv:2301.02111}, 2023.

\bibitem{le2023voicebox}
M.~Le, A.~Vyas, B.~Shi, B.~Karrer, L.~Sari, R.~Moritz, M.~Williamson, V.~Manohar, Y.~Adi, J.~Mahadeokar \emph{et~al.}, ``Voicebox: Text-guided multilingual universal speech generation at scale,'' \emph{Advances in neural information processing systems}, vol.~36, pp. 14\,005--14\,034, 2023.

\bibitem{ju2024naturalspeech}
Z.~Ju, Y.~Wang, K.~Shen, X.~Tan, D.~Xin, D.~Yang, Y.~Liu, Y.~Leng, K.~Song, S.~Tang \emph{et~al.}, ``Naturalspeech 3: zero-shot speech synthesis with factorized codec and diffusion models,'' in \emph{Proceedings of the 41st International Conference on Machine Learning}, 2024, pp. 22\,605--22\,623.

\bibitem{ren2024fewer}
Y.~Ren, T.~Wang, J.~Yi, L.~Xu, J.~Tao, C.~Y. Zhang, and J.~Zhou, ``Fewer-token neural speech codec with time-invariant codes,'' in \emph{ICASSP}.\hskip 1em plus 0.5em minus 0.4em\relax IEEE, 2024, pp. 12\,737--12\,741.

\bibitem{nagrani2020voxceleb}
A.~Nagrani, J.~S. Chung, W.~Xie, and A.~Zisserman, ``Voxceleb: Large-scale speaker verification in the wild,'' \emph{Computer Speech \& Language}, vol.~60, p. 101027, 2020.

\bibitem{mesaros2018multi}
A.~Mesaros, T.~Heittola, and T.~Virtanen, ``A multi-device dataset for urban acoustic scene classification,'' \emph{arXiv preprint arXiv:1807.09840}, 2018.

\bibitem{poria2019meld}
S.~Poria, D.~Hazarika, N.~Majumder, G.~Naik, E.~Cambria, and R.~Mihalcea, ``Meld: A multimodal multi-party dataset for emotion recognition in conversations,'' in \emph{ACL}, 2019.

\bibitem{ardila2020common}
R.~Ardila, M.~Branson, K.~Davis, M.~Kohler, J.~Meyer, M.~Henretty, R.~Morais, L.~Saunders, F.~Tyers, and G.~Weber, ``Common voice: A massively-multilingual speech corpus,'' in \emph{Proceedings of the twelfth language resources and evaluation conference}, 2020, pp. 4218--4222.

\bibitem{warden2018speech}
P.~Warden, ``Speech commands: A dataset for limited-vocabulary speech recognition,'' \emph{arXiv preprint arXiv:1804.03209}, 2018.

\bibitem{zen2019libritts}
H.~Zen, V.~Dang, R.~Clark, Y.~Zhang, R.~J. Weiss, Y.~Jia, Z.~Chen, and Y.~Wu, ``Libritts: A corpus derived from librispeech for text-to-speech,'' \emph{Interspeech}, 2019.

\bibitem{panayotov2015librispeech}
V.~Panayotov, G.~Chen, D.~Povey, and S.~Khudanpur, ``Librispeech: an asr corpus based on public domain audio books,'' in \emph{2015 IEEE international conference on acoustics, speech and signal processing (ICASSP)}.\hskip 1em plus 0.5em minus 0.4em\relax IEEE, 2015, pp. 5206--5210.

\bibitem{veaux2017cstr}
C.~Veaux, J.~Yamagishi, K.~MacDonald \emph{et~al.}, ``Cstr vctk corpus: English multi-speaker corpus for cstr voice cloning toolkit,'' \emph{University of Edinburgh. The Centre for Speech Technology Research (CSTR)}, vol.~6, p.~15, 2017.

\bibitem{he2024emilia}
H.~He, Z.~Shang, C.~Wang, X.~Li, Y.~Gu, H.~Hua, L.~Liu, C.~Yang, J.~Li, P.~Shi \emph{et~al.}, ``Emilia: An extensive, multilingual, and diverse speech dataset for large-scale speech generation,'' in \emph{Spoken Language Technology Workshop (SLT)}.\hskip 1em plus 0.5em minus 0.4em\relax IEEE, 2024, pp. 885--890.

\bibitem{chinen2020visqolv3opensource}
\BIBentryALTinterwordspacing
M.~Chinen, F.~S.~C. Lim, J.~Skoglund, N.~Gureev, F.~O'Gorman, and A.~Hines, ``Visqol v3: An open source production ready objective speech and audio metric,'' 2020. [Online]. Available: \url{https://arxiv.org/abs/2004.09584}
\BIBentrySTDinterwordspacing

\bibitem{941023}
A.~Rix, J.~Beerends, M.~Hollier, and A.~Hekstra, ``Perceptual evaluation of speech quality (pesq)-a new method for speech quality assessment of telephone networks and codecs,'' in \emph{ICASSP}, 2001.

\bibitem{5713237}
C.~H. Taal, R.~C. Hendriks, R.~Heusdens, and J.~Jensen, ``An algorithm for intelligibility prediction of time–frequency weighted noisy speech,'' \emph{IEEE Transactions on Audio, Speech, and Language Processing}, vol.~19, no.~7, pp. 2125--2136, 2011.

\bibitem{407206}
R.~Kubichek, ``Mel-cepstral distance measure for objective speech quality assessment,'' in \emph{Proceedings of IEEE Pacific Rim Conference on Communications Computers and Signal Processing}, vol.~1, 1993, pp. 125--128 vol.1.

\end{thebibliography}

\end{document}